\documentclass[conference]{IEEEtran}

\IEEEoverridecommandlockouts
\AtBeginDocument{%
  }

\usepackage{caption}
\usepackage{comment}
\usepackage{cite}
\usepackage{amsmath,amssymb,amsfonts}
\usepackage{algorithmic}
\usepackage{graphicx}
\usepackage{url}
\usepackage{tabularx}
\usepackage{textcomp}
\usepackage{xcolor}
\usepackage{float}
\usepackage{booktabs}
\usepackage{placeins}
\usepackage{multirow}
\def\eg{\emph{e.g., }}

\newcommand{\kp}[1]{\textcolor{teal}{#1}}
\newcommand{\cmh}[1]{\textcolor{red}{#1}}

\newcommand{\caf}{RoboCafé}
\newcommand{\cafe}{RoboCafé }

\definecolor{dihl}{RGB}{159,191,229}
\newcommand{\designimp}[1]{%
  \noindent\colorbox{dihl}{\bfseries\color{black}#1}%
  }
  
\begin{document}

\title{\cafe in the Open: Interaction Continuity in Long-Term Public Human-Robot Interaction}
\author{
\IEEEauthorblockN{
Kaitlynn Taylor Pineda$^{1,*}$, 
Kush Kumar Kushwaha$^{1,*}$, 
Jie Wang$^{1}$, 
Jiaming Du$^{1}$
}
\IEEEauthorblockN{
Anvii Mishra$^{1}$, 
Emilie Basu Suri$^{1}$, 
Angela Guo$^{1}$, 
Chien-Ming Huang$^{1}$
}
}


\twocolumn[{
  \renewcommand\twocolumn[1][]{#1}%
  \maketitle

  \begin{center}
    \includegraphics[width=\textwidth]{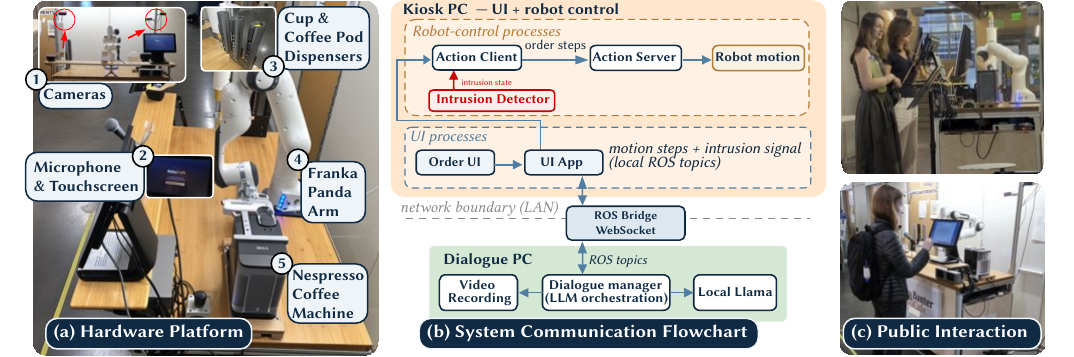}
    \captionof{figure}{\cafe\ Public Deployment Overview:
    a) hardware platform,
    b) programming \& communication flow, and
    c) kiosk interaction.}
    \label{fig:robocafe_deployment_overview}
  \end{center}
}]
\footnotetext[1]{%
Dept. of Computer Science,
Johns Hopkins University,
Baltimore, Maryland, USA.
{\ttfamily\small
\{kpineda3,kkushwa2,...\}@jhu.edu}
}

{%
\renewcommand{\thefootnote}{\fnsymbol{footnote}}%
\footnotetext[1]{%
These authors contributed equally to this work.
}%
}

{%
\renewcommand{\thefootnote}{}%
\footnotetext[0]{%
This work has been submitted to the IEEE for
possible publication. Copyright may be transferred
without notice, after which this version may
no longer be accessible.
}%
}

\begin{abstract}

As robots remain in public spaces over extended periods, they must maintain \emph{interaction continuity} by preserving and correctly applying context as people, encounters, and circumstances change.
To study interaction continuity in long-term public human-robot interactions, we developed \caf, an autonomous conversational coffee robot designed to support repeated interactions through task-aware dialogue, real-time multimodal perception, and memory of prior encounters. We deployed \cafe for 12 days in a university building, where it received 148 orders. The deployment involved repeat customers, passersby, changing groups, and back-to-back orders that repeatedly crossed the boundaries assumed by the system's order-centered interaction model.
We found that successful interaction continuity requires a robot to determine who is currently present, which prior context belongs to whom, where interactions begin and end, and whether its representation of an interaction matches what is occurring in the physical world. From these observations, we derive four system design requirements for maintaining interaction continuity in longitudinal public human-robot interactions: 
contextual interaction state,
persistent person grounding, 
explicit interaction life-cycle management, 
and
interaction observability.

\end{abstract}
\begin{IEEEkeywords}
Software Architecture for Robotic and Automation;
Long term Interaction; Social HRI; Service Robotics
\end{IEEEkeywords}

\section{Introduction}
\label{intro}
Robots are becoming increasingly capable of combining physical service tasks with open-ended conversation in public spaces. 
Recent advances in large language models (LLMs) make these interactions feasible by allowing robots to support more flexible, context-dependent dialogue rather than relying only on predefined conversational flows.
These robots may serve as receptionists, guides, assistants, or baristas,  
interacting not only with new people but also with people they have encountered in the past. 
As these interactions extend beyond one-time exchanges, robots have an opportunity to carry context across encounters. 
For example, a public service robot could
remember a returning customer, adapt to prior preferences, or continue a conversation from an earlier visit. 
This changes the systems problem from simply supporting a successful task interaction to maintaining coherent interaction context over time. 
We refer to this broader systems problem as \textbf{interaction continuity}: preserving and correctly applying context as people, encounters, and circumstances change.

Public-facing service robots provide a useful setting for researching this problem since task interactions are structured while the surrounding social interaction is more open-ended. 
Coffee service, for example, involves a repeated everyday task that can bring the same customers back to the robot over time, while also occurring in a shared social environment where people arrive together, interact around the service, and move through the space. 
An order thus provides a concrete task boundary around which a system can organize its behavior, but the social context surrounding that order does not necessarily begin and end at the same boundaries. 

In this work, we developed \caf, an autonomous LLM-powered conversational coffee robot designed for repeated public interaction (Fig. \ref{fig:robocafe_deployment_overview}). 
The system combined autonomous beverage preparation with task-aware dialogue, real-time multi-modal perception, and memory of prior customer encounters. 
We designed \cafe primarily around one-on-one interactions between the system and customers, with each coffee order providing a natural organizing unit for the interaction. 
We deployed the system for 12 days in a university building, where it received 148 orders. 
We use this deployment as a probe to study what systems require to support interaction continuity over repeated public encounters. 
From our deployment data, we derive four system design requirements:
\emph{contextual interaction state},
\emph{persistent person grounding},
\emph{explicit interaction life-cycle management},
and \emph{interaction observability}.
These requirements characterize the system capabilities needed to maintain coherent interaction context as people return and encounters change over time.

\section{Related Work}

\subsection{Modern Human-Robot Conversations}
Recent advances in LLMs have expanded robots’ conversational and reasoning capabilities \cite{zhang2023large, kim2024understanding}. 
This has enabled the incorporation of more natural, open-ended dialogue systems into robots for connection-building and rapport \cite{kim2024understanding, pineda2025see}.
However, when it comes to sustained periods of time,
LLM-based agents can struggle to maintain conversational coherence, continuity, and proper context across interactions that unfold over multiple sessions \cite{xu2021beyond, maharana-etal-2024-evaluating}. 
Similarly, prior studies of LLM-powered voice interfaces have documented recurring interaction breakdowns \cite{mahmood2025user}. 

\subsection{Interactions with Robots in Public Spaces}
Moving human-robot interaction (HRI) from controlled settings into public, real-world settings introduces conditions that are difficult to anticipate in development. 
At the system level, robots operating in industrial and service environments must contend with dynamic, under-specified, and potentially safety-critical conditions that may be vastly different from those encountered during development \cite{realworldphysical}.
Interactions are unscripted, participation is self-guided, and encounters with the robot may arise incidentally rather than through deliberate engagement\cite{rosenthal2020forgotten}.
Real users may leave conversations midway due to time constraints or boredom \cite{benyoussef2017dataset}.
Deployments ranging from museum guides \cite{1249720} and expo robots \cite{Siegwart2003RoboxAE} to shopping-mall assistants \cite{5557825}, an office receptionist \cite{lee2010receptionist}, and street-level trash-collecting robots \cite{bu2025making} show that passersby may engage in groups, play around with the robot's capabilities, and appropriate them in ways that extend beyond their intended roles. 
Public HRI is therefore shaped not only by a robot's designed behavior, but also by its surrounding social environment, its interpretability, and how people test and incorporate the robot into ongoing activity.

\subsection{Long-term and Repeated Human-Robot Interaction}
Research shows that interactions change across longer deployments. 
As initial interaction novelty fades, engagement dips or changes, and expectations about the roles of a robot may develop or shift over time \cite{leite2013social, Bickmore_Picard_2005, SEVERINSONEKLUNDH2003223, Horstmann2019}. 
These dynamics are more prominent in public deployments, where robots encounter both unfamiliar and returning users under uncontrolled conditions.
Long-term deployments in museums \cite{1249720}, shopping malls \cite{5557825}, reception settings \cite{gockley2005designing}, schools \cite{kanda2004interactive}, and food-service environments \cite{wilson2026faulty} have similarly exposed challenges in maintaining autonomy and supporting evolving user behavior. 
Over-time, users also tend to behave differently once the novelty of a deployed robot fades \cite{reimann2023social}.
However, most in-the-wild studies are largely centered on instances of isolated encounters or interactions aggregated across different users, with little work examining how the same individuals engage with a robot across repeated encounters.
Furthermore, much of this deployment literature work predates modern LLM-based conversation; it is unclear how robots with increasing conversational capabilities behave when embedded long-term in public spaces. 
Understanding how conversationally capable robots fare under sustained public use is therefore important for assessing both their readiness for deployment and their socially responsible integration into everyday environments. 

\subsection{Socially responsible HRI}
Operating in a new space among untrained members of the community requires the robot to act responsibly. 
Physical safety standards stem from industrial collaborative robotics \cite{ISO15066_2016}, but they must be adapted to a social context. Errors are an inevitable part of any deployed system, and people's responses may vary and be difficult to anticipate \cite{stiber2023using, stiber2026signal}.
Public robots may themselves also become targets of teasing, obstruction, and abuse \cite{Salvini2010HowSA, nomura2017why}.
Conversational robots powered by LLMs come with their own additional risks.
LLM-powered robots may be vulnerable to adversarial prompting \cite{robey2025jailbreaking} and privacy risks that arise from the sensitive information users disclose in a natural conversation \cite{Zhang_Jia_Lee_Yao_Das_Lerner_Wang_Li_2024}.
As social robots become increasingly conversational and embedded in public settings, 
understanding how their behavior influences the people and social environments around them
becomes central for socially responsible design and deployment.

\section{\cafe}
\label{depinfra}

While robotic coffee platforms are already being deployed in real-world public settings, 
these systems largely focus on automating beverage preparation and ordering within a domain where coffee consumption itself is often social.
This makes coffee service a useful testbed for examining what systems capabilities are needed as public-facing robots move beyond one-time interactions toward repeated conversational interaction.
Therefore,
we developed \cafe as a Deployable Research Product (DRP): 
a research-owned platform designed to be robust enough for extended real-world operation while remaining adaptable to different research questions \cite{matheus2026deployable}.
This section describes the deployment infrastructure that enabled \cafe
to operate consistently 
over 3 weeks in a high-traffic location, Monday---Friday from 11am to 6pm, 
following a 3-day public pilot the week before full-time deployment.
\begin{figure*}[!t]
    \centering
    \makebox[\textwidth][c]{%
        \includegraphics[width=\linewidth]{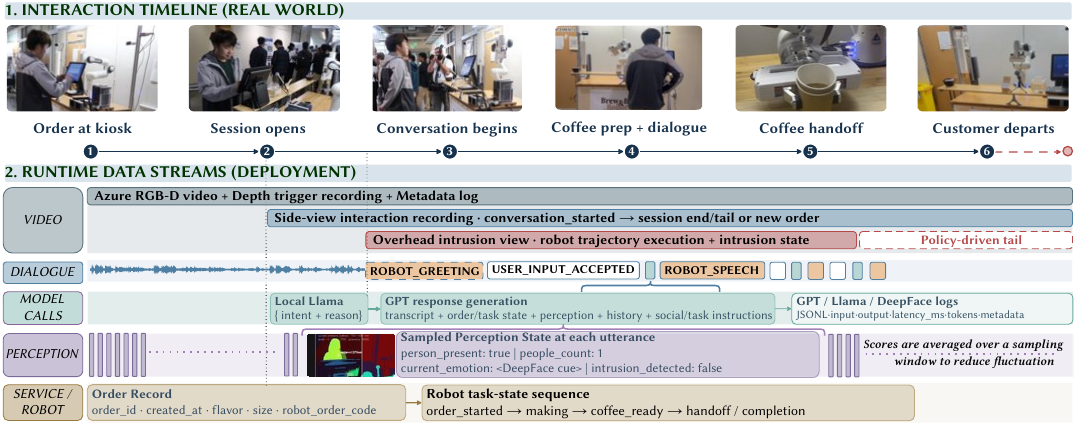}
    }
    \caption{
    \cafe interaction workflow overview. (1) User interaction timeline. (2) Alignment of recorded multimodal data.
    }
    \label{fig:interaction-timeline-data}
\end{figure*}

\textbf{Hardware.}
\label{depinfra:hd}
The \cafe is implemented as a compact integrated workstation for public-space deployment. 
A Franka robotic manipulator is mounted near the rear-center of the station, providing access to the commercial coffee machine on the customer-facing right side, the cup and capsule handling hardware on the left, and the beverage handoff region at the front. 
An elevated customer-facing platform supports the touchscreen ordering interface and pickup area, while speakers mounted below the interface provide audio output.

The workstation incorporates multiple cameras for perception and safety. 
Two RGB-D cameras are mounted at different heights on the rear-right structural 
rod
to observe users during interaction, while a separate safety camera monitors the robot workspace for any hand-reaching into workspace, which triggers a stop command to the robot.

A physical emergency-stop mechanism provides an additional means of immediately stopping operation. 

The platform is built around T-slot aluminum extrusions and a partially enclosed base containing computing hardware, power distribution, and cabling. 
Custom fixtures for capsule feeding, camera mounting, cup positioning, coffee-machine actuation, and peripheral support were added as enabling components for reliable long-term deployment. 

Fig.~\ref{fig:robocafe_deployment_overview} summarizes the integrated hardware and system architecture.

\textbf{Kiosk UI.}
\label{depinfra:ui}
The UI interface acts as the layer responsible for mediating the user's order selections with the underlying robotic system. 
It is a multi-page React frontend with a Node.js backend and a linear flow.
A landing page leads to a consent screen that displays the consent form,
requires the user to scroll through it, and collects a first name, 
last name, email address, 
and a three-way recording-consent choice (recording for research and publication, recording for research only, 
or no recording), with an optional checkbox permitting anonymous quotation.
Users who decline recording cannot order.
All others proceed to a menu of three drinks (medium roast, dark roast, and decaf) 
in two sizes, with at most four cups per order.
On submission, the backend appends one row per cup to the shared order sheet (timestamp, order ID, name, email, drink, and consent flags) and publishes the order to a ROS topic for the robot to work on.
The backend also subscribes to the robot's sub-step feedback;
when it observes the delivery sub-step of the last cup, it notifies the frontend over a websocket, which advances from a waiting page to a final page with a five-point experience rating that is written back to the same order rows.
The kiosk returns to the landing page after 30--60s of inactivity on any page, and can be remotely turned off.

\textbf{Network setup.}
\label{depinfra:sw:ns}
All components communicate over a wired network built around a central switch,
with the robot controller, robot PC, and dialogue PC on a dedicated VLAN.
CAT6 gigabit Ethernet is used throughout because wireless links dropped order packets and delayed cloud API calls during development.
ROS~1 topics are published between the two PCs, exchanging robot state and safety signals, while order and customer data uses a Google Sheet accessed every 3s.


\textbf{Robot Software.}
\label{depinfra:sw:rs}
The robot is programmed in ROS~1, and low-level control uses the \texttt{franka\_ros\_interface} and \texttt{panda\_robot} packages for waypoint based manipulation. An action client expands each order into a text based step list and sends the steps one at a time to an action server. This server executes the corresponding waypoint sequence once per cup. Then the current step and sub step details are published to the dialogue PC via the rosbridge server setup on the robot PC for operational context.
%
Workspace safety is handled by a separate ROS node that monitors an overhead USB camera, which detects hands using MediaPipe and publishes intrusion state for any overlap over the robot's workspace area. The action server subscribes to this state and keeps track of it using a separate thread to stop the motion within 0.2s smoothly and moves again on clearing of the state.
The intrusion node plays a recorded warning, and the dialogue PC interrupts any ongoing speech, speaks a warning in a distinct voice, and announces when the workspace is clear. This vision-defined boundary is informed by the guidance on collaborative operation in ISO/TS~15066 \cite{ISO15066_2016} and constrains the workspace without physical barriers. 

\textbf{Dialogue Processing.}
\label{depinfra:sw:mllm}
When an order is received, the dialogue PC opens the microphone, starts a Google Cloud streaming speech-recognition session, and proactively greets the customer. The session closes when the conversation ends. Speech output uses Google Cloud TTS. The microphone is muted while the robot speaks, preventing users from interrupting an utterance.

Each final transcript is classified by a local Llama-3.1-8B-Instruct model into sub-categories related to the topic of the user's utterance: coffee order status, the robot system, or small talk. 
Order-progress questions are answered directly from the active order and robot sub-step. 
All other turns are answered by GPT-4o-mini using a user intent recognition specific system prompt,
active order context, 
detected facial expression,
and the most recent 20 turns of the conversation.

\textbf{Perception and Interaction Memory.}
\label{depinfra:sw:mu}
An Azure Kinect above the kiosk drives presence sensing.
A depth background is captured at the start of each deployment day; foreground within a calibrated region in front of the kiosk indicates that a person is present, and a smaller trigger region starts video recording.
A MediaPipe person detector counts visible people, and an OpenCV Haar cascade with DeepFace's pretrained expression classifier \cite{parkhi2015deep} labels the largest visible face with one of seven expressions every fifth frame, averaged over a 15-frame window.
DeepFace is used only for expression; no face identity is computed.
This pipeline runs as a separate process and streams its state (presence, people count, and expression) to the dialogue controller over a local websocket.
When the recognizer completes an utterance, the controller pairs it with the latest perception state; a delayed or missing perception update does not block dialogue.

Presence gates the interaction: transcripts are ignored when no one is visible, 
the robot stops self-initiating when the customer steps away, 
and the conversation closes after a short grace period once the customer has left following the order-ready message.
The people count is logged with every turn but is not used to change behavior.
The expression label is the only perceptual cue included in the language-model prompt.

Continuity across visits is linked to the email address typed at the kiosk.
After each conversation, the local Llama model writes a summary record (order, topics discussed, and a short overview) to a per-email history file.
When a new order arrives, the controller looks up the lowercased email; if a history exists, it determines whether the customer returned the same day or on a later day and how many days have elapsed, and selects a returning-customer greeting accordingly.
The content of earlier conversations is stored but is not injected into later prompts as is, but converted into a summary, acting as a context for conversational memory.

\textbf{Logging.}
\label{depinfra:sw:log}
Every user and robot utterance is appended to a per-conversation JSONL log with start and end timestamps, the transcript, the intent label, the perception state at that moment (presence, people count, expression, and intrusion flag), and, for transcripts the system chose not to answer, the reason. Every model call is logged with its input, output, and latency. Three video streams are recorded: the Kinect view with audio, triggered by proximity; a second camera on the robot PC, triggered on conversation start and end; and the intrusion camera.

\textbf{Interaction Model and Design Scope.}
\label{depinfra:scope}
\cafe was designed for longitudinal interaction with one primary customer at a time.
The system binds each conversation to its ongoing order and the person who placed it, maintains the dialogue history within that conversation, grounds responses in the order and the robot's task state, incorporates live presence and expression, and stores per-customer summaries so that a returning customer is recognized and greeted as such.
Three capabilities were deliberately left outside this first deployment. 

\textit{Persistent biometric identity.}
Faces are processed for expression only with email collection for cross-visit identification. Retention of face features in public raised consent and data-storage concerns we chose not to take on, thus relying solely on user-reported information.

\textit{Co-presence as interaction membership.}
The system uses immediate presence to log people count without treating it as group membership since people captured in the camera's view could be customers, companions, or passersby.

\textit{Multiparty speaker attribution.}
The dialogue model assumes that speech accepted by the system comes from the customer. In the social condition, any utterance made while a face is visible is answered; in the task condition, a transcript-only gate estimates whether the utterance was addressed to the robot.
However, neither identifies who spoke. Focusing the first deployment on reliable primary-customer interaction let us observe, rather than pre-empt, how often public encounters exceed this model.

\section{\cafe \ as a Probe}
\label{probe}
As discussed in Section~\ref{depinfra}, the \cafe is a probe developed to publicly interact with the people ordering, like a barista, thus providing us insights into user behaviors and the social dynamics of conversation for service robots that will be deployed in the wild. In this section we define the procedures and metrics used to study interaction continuity in such systems.

\begin{table}[t]
    \centering
    \caption{Deployment overview by condition.
    \textsuperscript{\textdagger}Nine customers saw both (61 social-only, 38
    task-only, 9 both)}.
    \footnotesize
    \setlength{\tabcolsep}{4pt}
    \begin{tabular}{lccc}
        \toprule
        \textbf{Metric} & \textbf{Social} & \textbf{Task} & \textbf{All} \\
        \midrule
        Customers\textsuperscript{\textdagger} & 70 & 47 & 108 \\
        Orders & 86 (58.1\%) & 62 (41.9\%) & 148 \\
        Orders per day, \textit{Mdn} & 13.5 & 8 & 11 \\
        \quad IQR; range & 10.5--15; 10--24 & 8--11; 7--19 & 8--15; 7--24 \\
        Drinks served & 96 & 68 & 164 \\
        1-, 2-, 3-drink orders & 77; 8; 1 & 57; 4; 1 & 134; 12; 2 \\
        Order Window(sec), \textit{Mdn} & 73.2 & 170.5 & 171.6 \\
        \quad IQR & 168.3--178.7 & 166.2--173.1 & 167.6--176.4 \\
        \bottomrule
    \end{tabular}

    \label{tab:deployment-overview}
\end{table}

\subsection{Interaction Study Procedure}
\label{probe:procedure}
The main deployment ran from Monday to Friday 11AM to 6PM every weekday for 3 weeks.
The system only allowed for orders between 11AM and 6PM, after which the kiosk automatically deactivated.
Consent was collected through the kiosk. As described in Section~\ref{depinfra}, users were informed that interactions would be recorded, and consent to recording was required to place an order. Permission to reproduce quotes or video in publications was not required to use the system.

The deployment was initially designed as a 
$2$ (donation appeal) $\times$ $2$ (conversational style)
study, with donation appeal presented after the beverage making.
The robot framed its donation ask as supporting either (a) its continued operation or (b) a local charity. 
This donation request occurred after the customer experienced interacting with the robot that was engaged in one of two conversational style conditions. 

In the \textbf{social condition}, where the robot can initiate and respond to small talk, 
the robot generates engaging responses that acknowledge conversational and affective cues.
The robot may extend the social exchange through small talk or follow-up questions, and re-engages once after a period of silence.
In the \textbf{task condition}, responses are concise and order-focused, without follow-up questions.
The robot does not self-initiate conversation and a reply gate determines whether it responds. 
A pre-speech filter removes fillers and fragments from user utterances, after which GPT-4o-mini classifies the remaining utterance as either directed to the robot, not directed, or unclear. 
The robot system responds only to task-relevant, greeting, or utterances clearly directed to it. 
The social condition does not use this reply gate. Conditions were assigned by deployment and were switched approximately every two deployment days.
We initially planned to examine whether these appeals affected donation behavior across conversational conditions but dropped this analysis because too few donations were observed to support meaningful comparison, likely due to the campus setting and the financial status of a majority-student population. Hence, this paper focuses solely on interactions prior to the donation message.

\subsection{Analysis Definitions and Measures}
\label{probe:metrics}
Our primary unit of analysis is an \textbf{order}: one kiosk submission and its surrounding interaction.
The \textbf{customer} is the person who placed the order at the kiosk.
Because the deployed email-based identifier was unreliable, 
customer identity was reconstructed post hoc from video.
A \textbf{unique customer} is a video-verified identity with at least one coded order.
A \textbf{repeat customer} placed two or more coded orders, and a \textbf{return visit} is any order placed by a repeat customer after that customer's first order.

\textit{Encounter Composition.}
A \textbf{companion} is anyone the customer visibly interacted with during the order.
Each order was coded as a \textbf{solo} or \textbf{group} order.
An order was considered group when a companion joined within 80s of the robot's welcome message and remained until the order ended.
The 80s threshold corresponds to about 45\% of the 180s median order duration, 
so a qualifying companion was present for at least 55\% of the order.

\textit{Dialogue measures.}
We define the \textbf{order window} as the time from the robot's welcome message through its first order-ready message. 
Because the deployed system did not identify individual speakers, conversational measures describe detected transcripts rather than specific people.
For each order, we measured the number of
\textbf{accepted transcripts} to which the dialogue system generated a response, 
the number of words in accepted transcripts, 
and the number of 
\textbf{gated transcripts}, which the system received but chose not to answer. 
\textbf{Conversation participation} indicates whether an order contained at least one accepted transcript.

\textit{Analysis approach.}
Encounter composition and visit history 
were observed descriptively rather than experimentally assigned,
so we use these analyses to characterize the interaction patterns 
rather than make casual comparisons. 
We report counts, proportions, medians, and inter-quartile ranges.

\section{Deployment Observations}
\label{results}
Of 15 deployment days, three were excluded due to experimenter setup errors (\eg kiosk screen off, outdated build), 
leaving 12 days with six in each social and task condition.
Due to multiple exclusions, all social days fell during weeks 1 and 2, and all of week 3 ran the task condition.
The subsequent subsections report results observed.

\subsection{Deployment Context}
\label{results:context}
Table~\ref{tab:deployment-overview} summarizes the deployment across conditions.
Over the 12 days, the system served 164 coffees across 148 orders placed by 108 unique customers.
Orders were split between the social and task conditions, with 86 and 62 orders, respectively. 
The median number of orders per day was 11, with 13.5 on social days and 8 on task days. Most orders contained a single drink, with median order window of 171.6s.

\begin{table}[t]
    \centering
    \footnotesize
    \setlength{\tabcolsep}{4pt}
    \caption{Repeat customers and identity continuity.
    }
    \begin{tabular}{@{}llr@{}}
        \toprule
         & \textbf{Metric} & \textbf{Value} \\
        \midrule
        \multirow{5}{*}{\emph{Visits}}
          & Customers ordering once; twice or more & 93; 15 (13.9\%) \\
          & Orders by repeat customers & 55 / 148 (37.2\%) \\
          & Return-visit orders & 40 / 148 (27.0\%) \\
        \midrule
        \multirow{3}{*}{\emph{Kiosk entries}}
          & Junk or initials-only names & 41 / 148 (28\%) \\
          & Customers never typing a resolvable name & 21 (33 orders) \\
          & Repeat customers, new name every visit & 6 / 15 \\
        \midrule
        \multirow{4}{*}{\emph{Recognition}}
          & Recognized by email match & 16 (40\%) \\
          & \quad on a well-formed email & 11 (28\%) \\
          & Missed (fresh or mistyped email) & 24 (60\%) \\
          & False match (another person's email) & 1 \\
        \midrule
        \multirow{2}{*}{\emph{Conversation}}
          & Accepted user transcripts, Mdn & 6 vs.\ 0.5 \\
          & Visits with $\geq$1 accepted transcripts & 84.3\% vs.\ 50.0\% \\
        \bottomrule
        \label{tab:repeat_customers}
    \end{tabular}
\end{table}

\subsection{Repeated Encounters and Identity}
\label{results:repeat}
\begin{figure}[t]
    \centering
\includegraphics[width=\columnwidth]{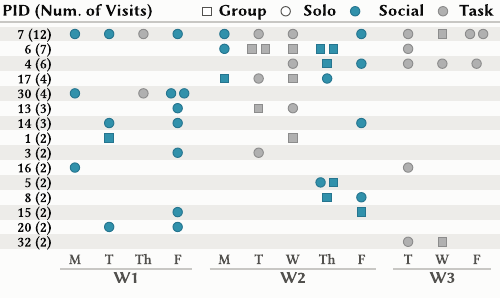}
    \caption{Repeat customer visits during the deployment.}
    \label{fig:robocafe_repeat}
\end{figure}

Table~\ref{tab:repeat_customers} summarizes these repeat customer encounters and identity continuity metrics.
Of the 108 unique customers, 15 placed two or more orders, accounting for 37.2\% of all orders; 27.0\% of orders were return visits.
However, the system correctly recognized only 16 of the 40 return visits from customer-provided identifiers, missing 24 and producing one false match.
Names were also inconsistent: 28\% of orders contained junk or initials-only names, 
21 customers never provided a resolvable name,
and 40\% of repeat customers used a different name across visits.
Conversation activity was lower on return visits than on first visits.
First visits had a median of six accepted transcripts, compared with 0.5 on return visits, and 84.3\% of first visits contained at least one accepted transcript compared with 50\% of return visits.
Return visits were unevenly distributed across customers, however: three customers accounted for 22 of the 40 return visits (Fig.~\ref{fig:robocafe_repeat}).

\subsection{Group Encounters}
\label{results:group}
Table~\ref{tab:group-solo-results} summarizes differences between solo and group encounters.
Of the 148 orders, 80 (54.1\%) were solo and 68 (45.9\%) were group. 
Thus, the majority of all users who experienced our system did so in a group setting. Although solo encounters outnumbered group encounters (80 vs. 68), each group encounter involved at least two people, so the 68 group encounters represented at least 136 people, compared with 80 users in solo encounters.
Company changed how much users spoke to the robot.
Group orders contained more 
accepted transcripts than solo orders (median 8 vs.\ 2),
more words in accepted transcripts 
(median 43 vs.\ 6), and more robot turns (median 10 vs.\ 3).
However, because the deployed dialogue system did not identify speakers, these transcripts cannot be attributed to the ordering customer. 
Group encounters therefore generated more speech where the system could not determine who was speaking.

\subsection{Deployment Cases}
\label{results:cases}

We detail four cases in which the customer's encounter with the system did not match the systems' interpretation.

\textit{Case A: A regular with no usable identifier}.
PID 7 visited on 12 days across both conditions (See Fig. \ref{fig:robocafe_repeat}) and typed a different gibberish string for each order name and email. 
Since the system would only identify a returning customer if they typed the same identifier, it
greeted them as a new customer each visit. 
Thus, summaries for this person accumulated under unrelated keys.

\textit{Case B: One group, five orders.} 
During W2-Wednesday around 1pm, five orders were placed in 16 minutes by five different people from one group. One person ordered once and appeared as a companion in three of the other orders. For each of the five people, the system opened a new conversation. The social encounter was longer than the order life-cycle, but the system did not represent interaction that extended beyond one order or spanned several orders back to back.

\textit{Case C: User adjusts system volume.}
On W3-Tuesday and Wednesday, PID 17 turned the tabletop speaker located under the bar counter off shortly after ordering and left it off. On the second day, they first told the robot to stop speaking and then turned the volume off as it replied.  It was observed that they first realized this accessibility when present with a group in an earlier interaction, where they fiddled with the volume knob.
This lead us to discard all interaction videos after their interaction on W3-Tuesday and Wednesday where the system volume remained off, including theirs.
The dialogue system continued to recognize speech, generate responses, and log robot turns spoken, under the assumption that its output reached the user.
Dialogue state and the physical state of the output channel diverged, and nothing in the system observed the difference.
\begin{table}[t]
    \centering
    \footnotesize
    \setlength{\tabcolsep}{2.5pt}
    \caption{Solo vs.\ group encounters.}
    \label{tab:group-solo-results}
    \begin{tabular*}{\columnwidth}{@{\extracolsep{\fill}}llrrr@{}}
        \toprule
         & \textbf{Metric} & \textbf{Solo} & \textbf{Group} & \textbf{All} \\
        \midrule
        \multirow{3}{*}{\emph{Composition}}
          & Orders & 80 (54\%) & 68 (46\%) & 148 \\
          & Identities & 55 & 62 & 108 \\
          & Customers, one mode only & 46 & 53 & 99 \\
        \midrule
        \multirow{5}{*}{\emph{Conversation}}
          & Accepted transcripts, Mdn & 2 & 8 & 4 \\
          & User words / transcript, Mdn & 6 & 43 & 16 \\
          & Robot turns, Mdn & 3 & 10 & 5 \\
          & Held a conversation & 65.0\% & 86.8\% & 75.0\% \\
          & Kept talking after ready & 20.0\% & 50.0\% & 33.6\% \\
        \bottomrule
    \end{tabular*}
\end{table}

\section{Discussion}
The \cafe deployment was designed around a bounded interaction model:
repeated conversational encounters with a primary customer, 
intersecting task and dialogue content, 
and prior conversational context stored for customers recognized across visits. 
Recognition relied on the email identifier associated with each order rather than continuous person tracking. 
Full multi-party dialogue was also outside the scope of this initial deployment.
In an open public space, however, this bounded interaction model becomes difficult to maintain.
Our deployment surfaced four challenges:
determining interaction membership, balanced persistent memory and privacy, robustness for open interactions, and interaction observability. 

\subsection{Interaction Membership is Dynamic}
A \designimp{contextual interaction state} is needed to determine who is participating in an interaction and how that participation changes over time. Our deployment showed that the set of people participating in an interaction cannot always be determined at the beginning of an order. 
Group orders accounted for almost half (46\%) of all orders, and we observed companions of customers
joining or leaving during an order, people remaining nearby between orders, 
and back-to-back interactions. Group interactions also produced substantially more accepted transcripts, words per transcript, and robot turns than solo interactions.
However, because the system could not attribute speech to individuals, these additional turns could not be reliably associated with the ordering customer (Section~\ref{results:group}).
This suggests that person detection alone is insufficient for establishing interaction membership.
A system may know that several people are present without knowing which people are participating, which person is speaking, or whether a newly detected person belongs to the ongoing interaction. Maintaining a contextual interaction state could allow future systems to track participant roles and changes in group membership over time, enabling them to distinguish an ordering customer from companions or passersby.

\subsection{Persistent IDs Needs More than Conversational Memory}
\designimp{Persistent person grounding} should be treated as a separate system capability from memory storage, allowing a robot to determine whether a current interaction belongs to a previously encountered person before applying information from prior encounters. Repeat encounters were a substantial part of the deployment. 
Fifteen customers returned to place multiple orders, accounting for over a third (37.2\%) of all orders, while return visits represented 27.0\% of orders. 
Additionally, identifiers were unusable in 60\% of repeat orders, 
and six of the fifteen repeat customers used different names across visits. 
Even the system's most frequent customer was unable to be identified on repeat visits, as shown in Section \ref{results:cases} (Case A). 
These observations show that continuity does not depend on conversational history alone, as context needs to be grounded to the correct person. 
However, stronger identity mechanisms introduce additional privacy and consent considerations, which were intentionally outside the scope of this deployment. Future systems should therefore treat persistent person grounding as an explicit design requirement, with separate consideration of identity reliability, privacy and consent, and which prior information should be associated with each person.

\subsection{Interactions Need Explicit Starts, Endings \& Transitions}
\designimp{Explicit interaction life-cycle management} should distinguish the progression of a social interaction from the boundaries of individual service tasks. The deployment demonstrated that an order is not necessarily equivalent to a social interaction. Section \ref{results:cases} (Case B) illustrates the limitation of using transactional boundaries as interaction boundaries.
While orders can represent distinct task episodes, they do not accurately map to continuing social encounters.
Future interaction managers should explicitly represent states such as start, continuation, transition, pause, and ending, allowing task-level state to be maintained while preserving the broader social interaction across service tasks.


\subsection{Observability Must Extend to the Interaction}
Finally, \designimp{interaction observability} should capture not only the internal state of individual system components, but also whether the system's actions remain meaningful in the physical interaction. The deployment revealed cases where the robot's internal state did not correspond to what was physically available to the user.
In Section \ref{results:cases} (Case C) 
where the user turned the volume off, 
the dialogue system had no awareness that its responses were no longer being delivered through the speaker. 
This suggests that observability for deployed HRI systems should extend beyond individual software and hardware components.
A system may report that perception, dialogue, and task execution are functioning while the interaction itself is no longer progressing as intended. 
Interaction-level awareness should therefore expose relevant state transitions and mismatches between system assumptions and the physical interaction, allowing researchers to determine not only what the system did, but whether those actions remained meaningful in context.

\section{Conclusion \& Future Work}

Overall, the \cafe deployment suggests that interaction continuity in longitudinal public human
robot interaction 
requires more than conversational memory.
Robots must maintain \textit{who is participating, 
which context belongs to whom, 
where interactions begin and end, 
and whether their internal state remains consistent with the physical encounter.}
These challenges motivate our four requirements:
contextual interaction state,
persistent person grounding, 
explicit interaction life-cycle management, 
and
interaction observability. 
These requirements become particularly important when robots move beyond controlled one-to-one interactions and operate repeatedly in open, social environments.
Many of these properties, including customer identity, interaction boundaries, and group composition, could not be recovered reliably from system logs and required retrospective, manual video analysis. 
Additional sensing such as speaker diarization would only partially address this problem, since separating voices does not establish which people belong to an interaction in an open public space. 
Future work should therefore explore privacy-conscious persistent identity, online interaction membership and boundary estimation, and architectures that make these states observable and recoverable when they fail.

\section*{Acknowledgment}
Authors used Claude (Anthropic) via Claude Code to assist with copy editing, table formatting, writing code for data pre-processing, and Fig. \ref{fig:robocafe_repeat} generation code.
All GenAI content was reviewed, verified, \& edited by the authors.



\bibliographystyle{IEEEtran}
\bibliography{references}



\end{document}